\documentclass[sigconf, nonacm]{acmart}

\usepackage{graphicx}
\usepackage{booktabs}
\acmYear{2024}

\usepackage{amsmath,amsfonts}

\usepackage{algorithm}
\usepackage{algorithmic}
\usepackage{hyperref}

\usepackage{tikz}
\usetikzlibrary{shapes,arrows}

\usepackage{numprint}

\iffalse
\newcommand{\memo}[1]{\todo[inline, color=green!40]{\textbf{Memo:} #1}}
\newcommand{\chris}[1]{\todo[inline, color=orange!40]{\textbf{Chris:} #1}}
\newcommand{\phat}[1]{\todo[inline, color=gray!40]{\textbf{Phat:} #1}}
\newcommand{\apo}[1]{\todo[inline, color=red!15]{\textbf{Apo:} #1}}
\newcommand{\tommy}[1]{\todo[inline, color=blue!15]{\textbf{Tommy:} #1}}
\else
\newcommand{\memo}[1]{}
\newcommand{\chris}[1]{}
\newcommand{\phat}[1]{}
\newcommand{\apo}[1]{}
\newcommand{\tommy}[1]{}
\fi

\begin{document}
\title{One Noise to Rule Them All: Multi-View Adversarial Attacks with Universal Perturbation}

%
%

\author{Mehmet Ergezer}
\orcid{0000-0001-6627-3667}
\affiliation{%
  \institution{Wentworth Institute of Technology}
  \streetaddress{550 Huntington Ave}
  \city{Boston}
  \state{MA}
  \country{USA}
  \postcode{}
}
\email{ergezerm@wit.edu}

\author{Phat Duong}
\affiliation{%
  \institution{Wentworth Institute of Technology}
  \streetaddress{550 Huntington Ave}
  \city{Boston}
  \state{MA}
  \country{USA}
  \postcode{}
}
\email{duongp@wit.edu}

\author{Christian Green}
\affiliation{%
  \institution{Wentworth Institute of Technology}
  \streetaddress{550 Huntington Ave}
  \city{Boston}
  \state{MA}
  \country{USA}
  \postcode{}
}
\email{greenc10@wit.edu}

\author{Tommy Nguyen}
\affiliation{%
  \institution{Wentworth Institute of Technology}
  \streetaddress{550 Huntington Ave}
  \city{Boston}
  \state{MA}
  \country{USA}
  \postcode{}
}
\email{nguyent68@wit.edu}

\author{Abdurrahman Zeybey}
\affiliation{%
  \institution{Wentworth Institute of Technology}
  \streetaddress{550 Huntington Ave}
  \city{Boston}
  \state{MA}
  \country{USA}
  \postcode{}
}
\email{zeybeya@wit.edu}


\begin{abstract}

This paper presents a novel ``universal perturbation'' method for generating robust multi-view adversarial examples in 3D object recognition. Unlike conventional attacks limited to single views, our approach operates on multiple 2D images, offering a practical and scalable solution for enhancing model scalability and robustness. This generalizable method bridges the gap between 2D perturbations and 3D-like attack capabilities, making it suitable for real-world applications.

Existing adversarial attacks may become ineffective when images undergo transformations like changes in lighting, camera position, or natural deformations. We address this challenge by crafting a single universal noise perturbation applicable to various object views. Experiments on diverse rendered 3D objects demonstrate the effectiveness of our approach. The universal perturbation successfully identified a single adversarial noise for each given set of 3D object renders from multiple poses and viewpoints. Compared to single-view attacks, our universal attacks lower classification confidence across multiple viewing angles, especially at low noise levels. A sample implementation is made available
\footnote{\href{https://github.com/memoatwit/UniversalPerturbation}{https://github.com/memoatwit/UniversalPerturbation}}.

\end{abstract}


\maketitle              

%
%
%









\section{Introduction}




In recent years, deep learning has revolutionized various fields, including 3D object recognition \cite{klokov2017escape, qi2017pointnet, le2018pointgrid, wang2019normalnet, zhi2018toward}, where it plays a crucial role in applications like augmented reality \cite{liu2019edge, apicharttrisorn2019frugal, li2020object}, autonomous driving \cite{li2019stereo, ravindran2020multi,zhang2020dnn}, and robotics \cite{hu20193, hossain2016object, finn2015learning}. However, a major challenge lies in the vulnerability of deep learning models to adversarial attacks \cite{ren2020adversarial}. These attacks entail generating imperceptible noise that, when added to a model's input, leads the model to make incorrect predictions. This poses serious security and safety concerns, particularly in critical applications like self-driving cars where misclassification can have catastrophic consequences \cite{goodfellow2015explaining, luo2022entangling}. Ensuring adversarial robustness, the ability of models to resist such attacks is crucial for real-world deployments.

Conventional adversarial attacks primarily focus on crafting perturbations for single views of objects, exploiting limitations in 2D image understanding \cite{szegedy2013intriguing, madry2017towards}. These attacks face limitations when applied to 3D objects, as they often fail to transfer across different viewing angles and lack robustness to real-world variations in perspective. This motivates the need for multi-view adversarial attacks, capable of fooling object recognition models across diverse poses and viewpoints.

In this work, we propose a novel ``universal perturbation'' method for generating robust multi-view adversarial images. Our approach departs from traditional per-view attacks by crafting a single noise perturbation applicable to various views of the same object. This single-noise, multi-view attack offers several advantages:

\textbf{Robustness:} The trained noise exhibits improved transferability across diverse viewing angles, leading to more robust attacks compared to single-view methods.

\textbf{Efficiency:} Training a single perturbation is faster and less computationally intensive compared to generating individual perturbations for each view.

\textbf{Scalability:} Our approach is readily applicable to different object categories and viewpoints, offering a generalized framework for multi-view adversarial attacks.

This paper investigates the effectiveness of our ``universal perturbation'' method in comparison to conventional single-view attacks. We conduct comprehensive experiments on various 3D object datasets, evaluating the attack success rates, and transferability across different views
We believe this work represents a significant step towards developing efficient and robust adversarial attacks for 3D object recognition, paving the way for improved model security and robustness in real-world applications. 

The paper is organized as follows: Section~\ref{sec:related} introduces related adversarial attack methods, focusing on single-view attacks. The universal perturbation algorithm is formulated in Section~\ref{sec:universal}. Sections~\ref{sec:experiments} and ~\ref{sec:results} present the setup for our experiments, including object selection and rendering, adversarial attacks, related metrics, and results. Section~\ref{sec:Discussion} discusses the potential and limitations of universal perturbation.  Section~\ref{sec:conclusion} provides conclusion remarks. 



\section{Related Work}\label{sec:related}

In this section, we provide an overview of the existing adversarial algorithms that are designed for single-view attacks.  




\subsection{Fast Gradient Sign Method}
One of the earliest adversarial attack methods, Fast Gradient Sign Method (FGSM), was proposed in 2014 \cite{goodfellow2015explaining}. The method was popular for forging adversarial examples due to its simplicity in generation and implementation and relatively low cost of execution. FGSM attacks rely heavily on knowing the architecture and the weight of the victim model---known as a white-box attack---which in turn may make the method poor when it comes to transferability and in black-box settings where only the input features are known.

FGSM works by first calculating the loss based on the predicted class after performing forward propagation. We determine the gradients with respect to the input image and update the image's pixels in the direction that maximizes the loss. In  Equation~\ref{eq:fgsm}, we let $X$ and $y$ be the input image and true label respectively. We assume $X$ as a 3-D matrix (width × height × color). \(\epsilon\) represents a constant that determines the strength of the perturbation. \(\nabla_X J(X, y_{true})\) is the gradient of the model's loss with respect to \(X\). The perturbation is calculated by taking the sign of this gradient and adding it to the original image. We initialize $X^{adv}_0 = X$.

\begin{equation}
X^{adv} = X + \epsilon  \cdot\text{sign}( \nabla_X J(X, y_{true}))   
\end{equation}\label{eq:fgsm}

While in this study we focus primarily on untargeted adversarial attacks, the FGSM algorithm can be updated to target specific class labels. This can be accomplished by simply changing the direction of the gradients to minimize the loss between the targeted and predicted labels as shown in Equation~\ref{eq:fgsmtar}.

\begin{equation}
X^{adv} = X + \epsilon \cdot \text{sign}( \nabla_X J(X, y_{target}))  
\end{equation}\label{eq:fgsmtar}

\subsection{Basic Iterative Method}
Kurakin et al. proposed several new iterative versions of FGSM, including the Basic Iterative Method (BIM) and iterative least-likely class method (ILCM) \cite{kurakin2017adversarial}. Both of these methods brought adversarial attacks to the physical world by inputting captures from a mobile phone camera, instead of feeding source images directly to the AI model. ILCM extends BIM by allowing the attacker to specify a desired target label for the attacked object. 


The Basic Iterative Method (BIM)  extends the idea of FGSM by applying it iteratively, allowing for a more fine-tuned manipulation of the input image. This iterative approach often results in more effective adversarial examples than single-step perturbation. Note that intermediate results are clipped at a pixel level after each step. Just like with FGSM, we initialize BIM, $X^{adv}_0 = X$. $N$ in Equation~\ref{eq:bim} represents the iteration count. 

\begin{equation}
X_{N+1}^{adv} = \text{Clip}_{X,\epsilon} \{X_{N}^{adv} + \epsilon  \cdot\text{sign}( \nabla_X J(X_{N}^{adv}, y_{true}))\}
\end{equation}\label{eq:bim}

\subsection{Other Notable Adversarial Algorithms}

Another prevalent method is the Carlini and Wagner attack
\cite{carlini2017evaluating}. Its goal is to generate a perturbation to the input data that causes the target neural network to misclassify while limiting the disturbance's perceptibility. The attack consists of solving an optimization problem with numerous objectives and constraints.

However, due to low success rates with adversarial attacks in black-box settings, Dong et al. proposed a class of momentum-based iterative algorithms to boost adversarial attacks called MI-FGSM \cite{dong2018boosting}. This class of adversarial attacks can be used for white-box attacks as well as outperformed other one-step gradient methods in a black-box setting.

This still has not addressed the transferability issue with black-box-based FGSM attacks, which is where  \cite{lin2020nesterov} proposed two new methods in 2019, Nesterov Iterative Fast Gradient Sign Method (NI-FGSM), and Scale-Invariant Attack (SIM).

Since then, various other modifications have been made to improve adversarial attacks. Notably, in 2020, \cite{andriushchenko2020understanding} proposed a regularization method named GradAlign, which prevents catastrophic overfitting that is common in FGSM models.

\section{Universal Perturbation}\label{sec:universal}

The methodologies outlined in Section~\ref{sec:related} are all tailored to attack a solitary image, denoted as $X$, generating a corresponding adversarial noise, $X^{adv}$, for each individual input. However, our objective is to extend single-view attacks; we aim to devise a singular perturbation that can be universally applied across various perspectives of the same object or even different objects, $\mathbb{X}^{\text{adv}}$. 

To achieve this, we propose a modification to the basic iterative method. Instead of computing gradients with respect to the input image, we consider the gradients with respect to the adversarial noise itself. This adjustment decouples the number of input images from the shape of the generated adversarial noise. Consequently, it allows the attacker to control both the shape of the input images and the associated noise, independently. This update allows simultaneous input of multiple views of the same object, enabling the calculation of a solitary adversarial noise capable of collectively compromising the recognition of all these distinct perspectives at once.






Equation~\ref{eq:universal} introduces the universal perturbation calculation where $X$ is stacked input images as a 4-D tensor ( image count × width × height × color), $\mathbb{X}$ is the calculated perturbation as a 3-D matrix (width × height × color), matching the shape of an individual image in $X$.  $N$ is the number of desired iterations as defined in BIM, costing calls to the backpropagation. $\epsilon$ is a hyperparameter controlling the scale of the attack as used in FGSM. \( \nabla_{\mathbb{X}_{N}} J(\mathbb{X}_{N}^{\text{adv}}, y_{\text{true}}) \) is the gradient of the cross-entropy loss function of the model with respect to the calculated perturbation.

\begin{equation}
\begin{aligned}
\mathbb{X}_{N+1}^{\text{adv}} &= \text{Clip}_{X,\epsilon} \{\mathbb{X}_{N}^{\text{adv}} + \epsilon \cdot \text{sign}(\nabla_{\mathbb{X}_{N}} J(\mathbb{X}_{N}^{\text{adv}}, y_{\text{true}}))\}
\end{aligned} \label{eq:universal}
\end{equation}

Unlike BIM, we initialize $\mathbb{X}^{adv}_0 = X + r$ where $r\sim U(-0.01, 0.01)$. Section~\ref{sec:Discussion} discusses the reasoning and intricacies associated with this initialization.




We make our implementation of the universal perturbation publicly available
\footnote{\href{https://github.com/memoatwit/UniversalPerturbation}{https://github.com/memoatwit/UniversalPerturbation}} and present its pseudocode at \ref{alg:universal_attack}.

\begin{algorithm}
\caption{Generate Universal Perturbation}
\textbf{Input}: images of object(s) from different views\\
\textbf{Parameter}: target\_model, $\epsilon=0.01$, num\_iterations=1, regularization=1., clip\_value=1.\\
\textbf{Output}: Adversarial noise\\
\begin{algorithmic}[1]
    \STATE $perturbation \gets \text{random.uniform(shape=images[0].shape, range$=\pm0.01$}$)
    \FOR{$it$ in range(num\_iterations)}
        \STATE \quad \text{Compute gradient($perturbation$)}
        \STATE \quad \quad $adv\_images \gets images + perturbation$
        \STATE \quad \quad $predictions \gets target\_model(adv\_images)$ 
        \STATE \quad \quad $loss \gets \text{crossentropy(perturbation)}$
        
        \STATE \quad \quad $loss \mathrel{+}= regularization \cdot \text{norm}(perturbation)$ 

        \STATE \quad \quad$gradients \gets \text{gradient}(loss, perturbation)$
        \STATE \quad \quad $gradients \gets \text{clip}(gradients, clip\_value)$
        \STATE \quad \quad$perturbation \mathrel{+}= (\epsilon \cdot \text{sign}(gradients))$
        \STATE \quad \quad$perturbation \gets \text{clip}(perturbation, clip\_value)$

    \ENDFOR
    \STATE \textbf{return} $perturbation$
\end{algorithmic}
\label{alg:universal_attack}
\end{algorithm}



\section{Experiments and Results}\label{sec:experiments}

In this section, we describe the system setup employed to automatically generate multi-view renders of objects, how we evaluate the FGSM, BIM, and universal perturbation attacks, and our analysis of their results. 

The experiments detailed in the following sections will substantiate our approach by assessing its true class accuracy drop, transferability across angles, and noise imperceptibility, while comparing the outcomes with those obtained using FGSM and BIM. Additionally, we will showcase the computational efficiency of our model in comparison to FGSM and BIM for generating transferable multi-angle noise.









\subsection{Experimental Setup}

To showcase the effectiveness of our model in generating robust noise across multiple angles, we conducted experiments comparing our universal perturbation method with common adversarial attack methods: FGSM and BIM. The objective was to identify the optimal noise level that significantly reduces a classifier's confidence in the object's true class while demonstrating that our universal perturbation method produces more robust noise than FGSM or BIM from multiple viewpoints of an object.

Algorithm \ref{alg:system} provides an overview of our experimental process. Unlike FGSM and BIM, our universal perturbation approach generates a single noise pattern applicable to all angles, resulting in consistent adversarial outcomes. 
Our universal perturbation method is designed to be more resilient in producing multi-angle consistent noise, enabling attacks in 3D space. To evaluate our method alongside FGSM and BIM, we conducted multiple runs of FGSM and BIM to generate noise for various angles of our objects.

\begin{algorithm}[tb]
\caption{Experimental System}
\label{alg:algorithm}
\textbf{Input}: 3D Blender model with texture\\
\textbf{Parameter}: $\epsilon$, $n\_angles$ to render, num iterations\\
\textbf{Output}: Adversarial image and its classification\\
\begin{algorithmic}[1] 
\STATE Render object from multiple angles, $n\_angles$, in Blender
\STATE Split $n\_angles$ images to two: $tr_n$ to generate noise and $ts_n$ to test the generate noise(s)
\STATE Identify universal perturbation for $tr_n$ views to find a single $perturbation$  for all $tr_n$
\STATE Generate adversarial images: $ts_n^{adv} \gets ts_n + (\epsilon \cdot perturbation )$
\STATE Classify the adversarial $ts_n^{adv}$ images with MobileNetv2

\STATE Repeat above for FGSM and BIM, except each $tr_n$ gets a unique $perturbation$ calculated.


\STATE \textbf{return} top-1 and top-5 accuracy for these attacks 
\end{algorithmic}
\label{alg:system}
\end{algorithm}

\subsection{3D Model Creation and Rendering in Blender from Multiple Views.}


To generate diverse images of objects, we employed Blender and various 3D models. For each object, we rendered images from ten distinct view angles, ensuring consistent recognition by our classification model across perspectives. 
Camera angles were randomly selected based on a spherical coordinate system centered on the object, incorporating a 15\% random deviation for robustness. This mimics natural variations in viewing angles encountered in real-world scenarios. Constant lighting position relative to the camera prevented shadowed areas that could hinder recognition. Additionally, we utilized five diverse 3D models that were categorized by imagenet \cite{deng2009imagenet} as discussed in Section~\ref{sec:objsel}.

\subsection{Object Selection}\label{sec:objsel}

To validate our methods, we selected five diverse 3D objects: baseball \cite{baseball_cite},
dining table \cite{table_cite}, 
lemon \cite{lemon_cite},  
shovel \cite{shovel_cite}, 
and tractor \cite{tractor_cite}, 
that were readily renderable on Blender from various viewpoints. These objects offer varying shapes, sizes, and textures, representing a range of potential real-world applications. 

Our pipeline utilizes Blender to generate multi-angle views of objects, at 224 by 224 resolution,  suitable for our image classification model, MobileNetV2. Currently, we render ten images of each object from different angles. These angles are determined using sinusoidal and cosinusoidal functions, ensuring even distribution across the viewing sphere. Additionally, random rotations are introduced to simulate natural object orientations.




In each case, we generate noise based on five training images and then apply the noise to five testing images to evaluate robustness. For the universal attack, a single noise is generated based on the five training images and then applied to the remaining images in the testing set. In FGSM and BIM, we generate five different noises for each image in the training set and then apply these noises to five separate testing images. The resulting noisy images from each model are then passed through MobileNetV2, from which we extract the top-1 and top-5 predicted classes.





\section{Results of Multi-view Attacks}\label{sec:results}

\subsection{Accuracies and Confidences}
Table~\ref{tab:combined_accuracy_comparison} summarizes the average accuracies of predicting the true object label when images are subjected to different adversarial attacks. The comparison encompasses three attack methods: FGSM, BIM, and our proposed universal perturbation method. We split the table in two to analyze the robustness of these attacks across 25 training and  25 unseen images corresponding to multiple poses of our five objects. 

For the test images, the best attack result for each $\epsilon$, corresponding to the lowest probability of true label, is bolded. We note that the universal attacks provide the highest success at the lowest noise levels, while FGSM is more successful at higher $\epsilon$. We also list the mean and standard deviation of true label prediction rates where a lower values indicate a more successful attack. Universal perturbation provides the minimum average true level probability across all $\epsilon$. This indicates the effectiveness of the proposed algorithm on unseen images at the most challenging  $\epsilon$ levels. 

Table~\ref{tab:combined_accuracy_comparison} also shows the effectiveness of the FGSM and universal attacks on training images,  achieving success even at lower epsilon values. This suggests that during the training phase, the model may become overly adapted to these perturbations, a vulnerability that these attacks exploit efficiently. As  $\epsilon$ increases to 15, these attacks also demonstrate marked success on the test images, with accuracies plummeting to below 5\%. This indicates that at the $\epsilon$ threshold of 15, the adversarial modifications are sufficiently pronounced to compromise the model's ability to generalize, leading to a significant degradation in performance on unseen data.

\begin{table}[ht]
\centering
\label{tab:combined_accuracy_comparison}

\begin{tabular}{|l||ccc||ccc|}
\hline
\multicolumn{1}{|c||}{$\epsilon$} & \multicolumn{3}{c||}{Train Images} & \multicolumn{3}{c|}{Test Images} \\ \cline{2-7} 
            & \phantom{ }FGSM    & BIM      & Universal\phantom{ }   & \phantom{ }FGSM    & BIM      & Universal \phantom{ } \\ \hline
        0.00 & 0.70 & 0.70 & 0.70 & \textbf{0.65} & \textbf{0.65} & \textbf{0.65} \\ 
        0.50 & 0.13 & 0.65 & 0.30 & 0.56 & 0.57 & \textbf{0.47} \\ 
        1.00 & 0.11 & 0.62 & 0.27 & 0.62 & 0.56 & \textbf{0.51} \\ 
        3.00 & 0.18 & 0.62 & 0.35 & 0.52 & 0.61 & \textbf{0.49} \\ 
        5.00 & 0.19 & 0.57 & 0.34 & \textbf{0.36} & 0.56 & 0.38 \\ 
        10.00 & 0.21 & 0.39 & 0.28 & \textbf{0.19} & 0.43 & 0.31 \\ 
        15.00 & 0.08 & 0.35 & 0.09 & \textbf{0.04} & 0.32 & 0.09 \\ 
        30.00 & 0.00 & 0.06 & 0.00 & \textbf{0.00} & 0.02 & \textbf{0.00}\\ 
        50.00  & 0.00 & 0.00 & 0.00 & \textbf{0.00} & \textbf{0.00} & \textbf{0.00} \\                           
 \hline
Mean & 0.18  & 0.44	 & 0.26	 & 0.33	 & 0.41	 & \textbf{0.32} \\
Std $(\pm)$ & 0.21	 & 0.26	 & 0.22	 & 0.27	 & 0.25	 & \textbf{0.24} \\
 \hline

\end{tabular}
\caption{Average accuracy of MobileNetV2 under different adversarial attacks at various epsilon values for both train and test images. $\epsilon=0.00$ indicates clean images without an attack.}
\end{table}

\begin{figure}[h]
    \centering
    \includegraphics[width=\linewidth]{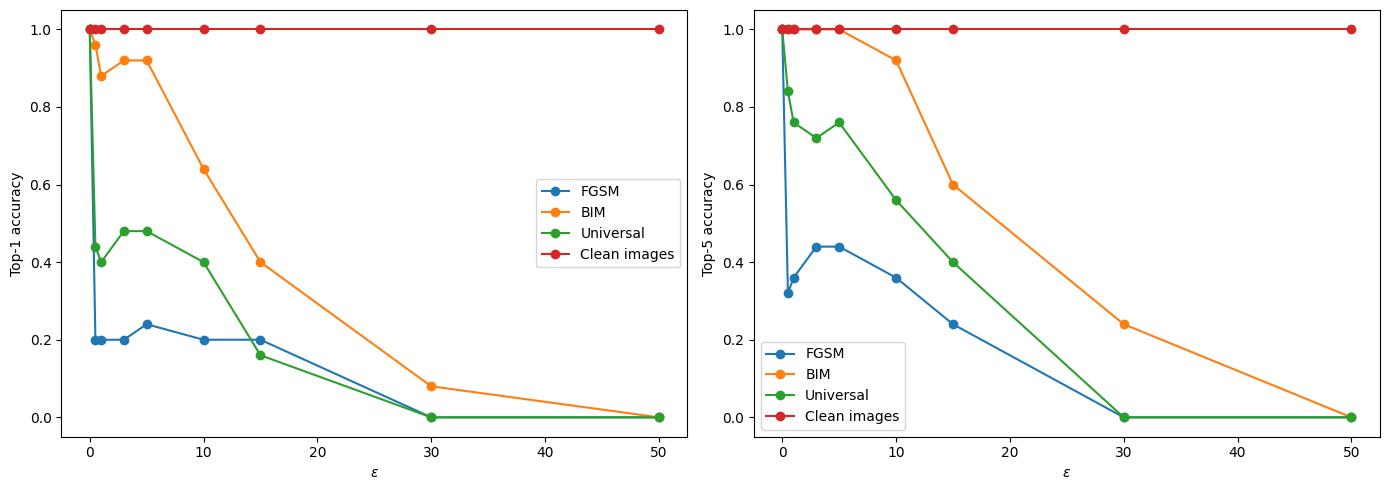}
    \caption{Top-1 and top-5 accuracies of MobileNetV2, after adversarial attacks with \(\epsilon\) values ranging from 0.5 to 50, were compared with those for clean Images—unmodified images from the dataset. The accuracies were calculated using a set of 25 \textbf{train} images, which were rendered from 5 distinct 3D object models.}
    \label{fig:PlotTopTrain}
\end{figure}
\begin{figure}[h]
    \centering
    \includegraphics[width=\linewidth]{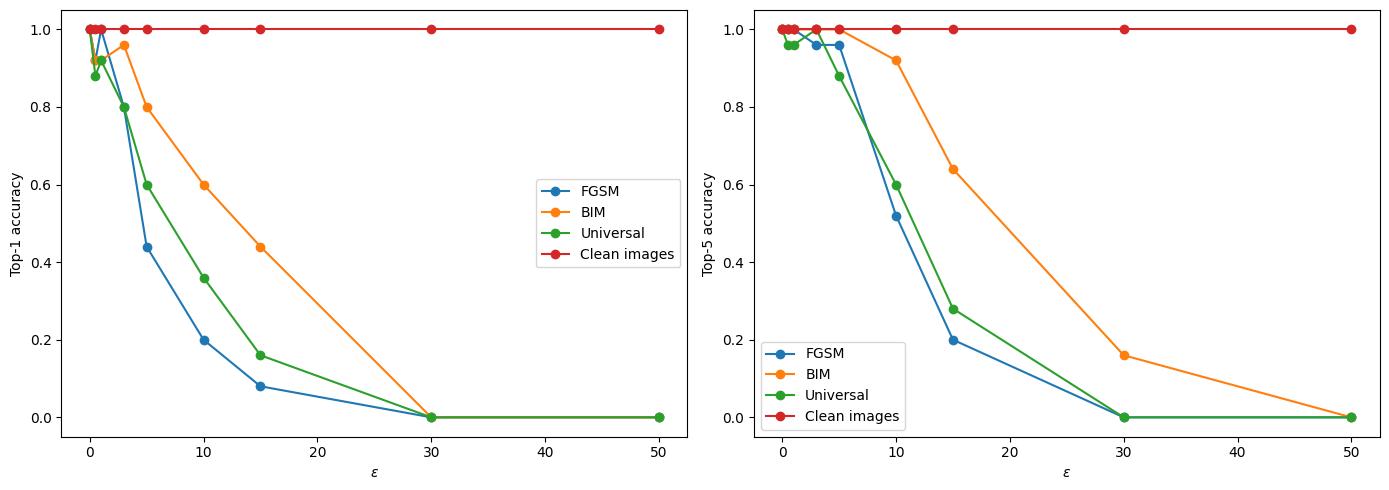}
    \caption{Top-1 and top-5 accuracies of MobileNetV2, after adversarial attacks with epsilon \(\epsilon\) values ranging from 0.5 to 50, were compared with those for clean images—unmodified images from our dataset. The accuracies were calculated using a set of 25 \textbf{test} images, which were rendered from 5 distinct 3D object models.}
    \label{fig:PlotTopTest}
\end{figure}
\begin{figure}[h]
    \centering
    \includegraphics[trim={0 0 0 0cm},clip, width=0.4\textwidth]{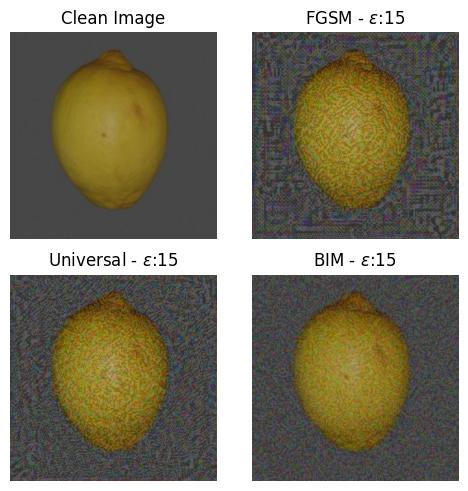}
    \caption{Comparison of adversarial attacks on the lemon object, each with an \(\epsilon\) value of 15. MobileNetV2 misclassifies these adversarial images in both top-1 and top-5 predictions. The noise level generated by BIM appears smoother than the other methods.}
    \label{fig:epsilon_comp}
\end{figure}

In Fig.~\ref{fig:PlotTopTrain} and Fig.~\ref{fig:PlotTopTest}, we evaluate the top-1 and top-5 classification accuracies for training and test images, respectively. Sample images of our lemon object, perturbed at $\epsilon = 15$ or a 15\% noise level, are shown in Fig.\ref{fig:epsilon_comp} to demonstrate the different noise types generated by each algorithm. We observe that as noise level, $\epsilon$ increases, we achieve successful misclassifications across all models. Both the FGSM and universal attacks exhibited faster degradation in top-1 and top-5 accuracies compared to BIM. 


Based on the results, it is evident that our universal perturbation method outperforms BIM, as indicated by consistently lower accuracy levels for the true class across all images of the object displayed. While our method may not always achieve the same level of performance as FGSM, it's essential to recognize that our approach aims to generate a single noise applicable across multiple angles of an object. In contrast, FGSM and iterative methods generate separate noises designed for each angle of the object individually.

\subsection{Adversarial Object Samples}
While we evaluated our method on various objects, here we analyze the shovel due to its intricate shape and sensitivity to viewing angle. This sensitivity makes it an excellent candidate for showcasing the effectiveness and robustness of our universal perturbation noise. 

\begin{figure}[h]
    \centering
    \includegraphics[trim={0 0 0 9cm},clip, width=0.5\textwidth]{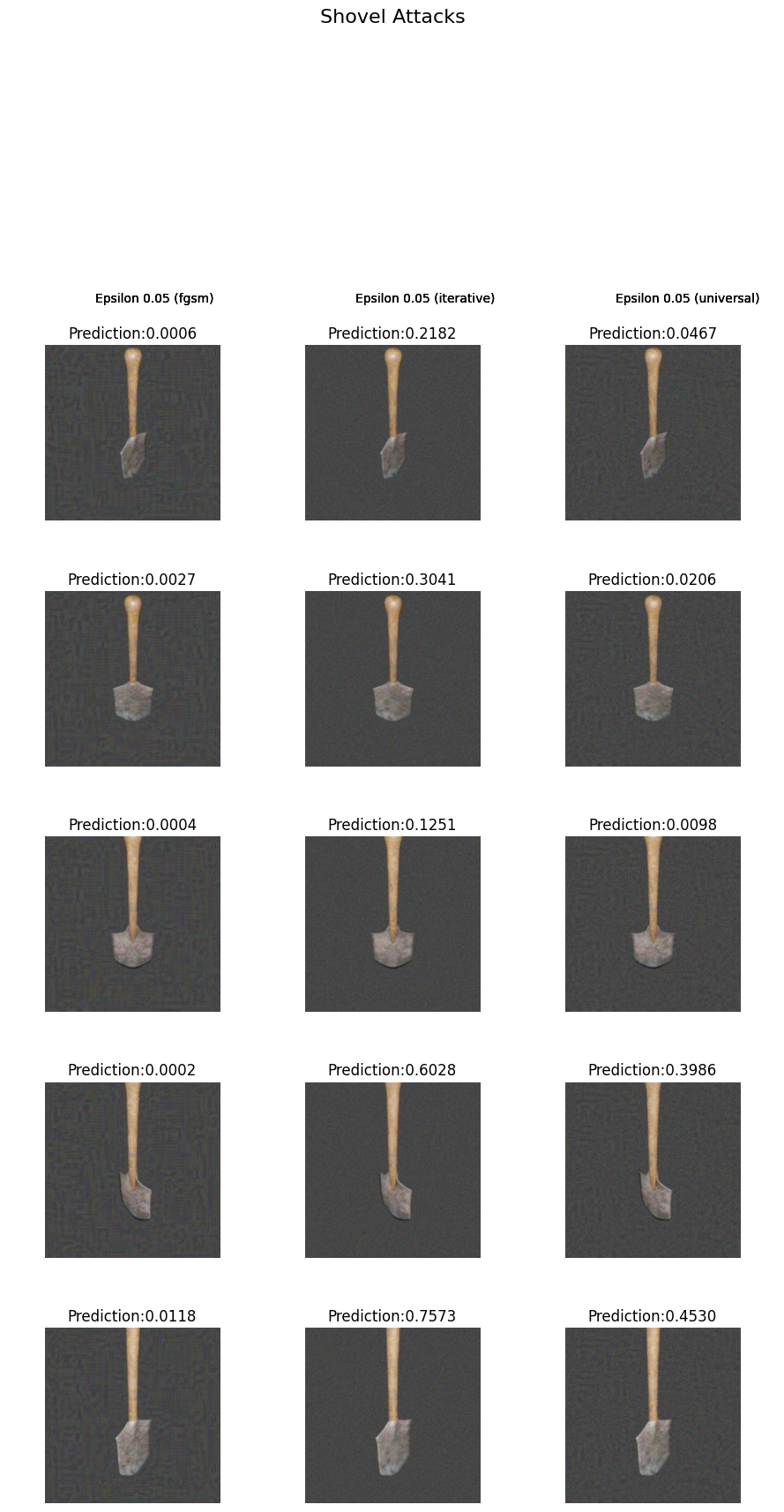}
    \caption{Training images for the shovel object as attacked by FGSM, BIM, and Universal algorithms, respectively, at $\epsilon=0.05$. The prediction confidence of the correct label for each image is indicated on top.}
    \label{fig:ShovelAttacks}
\end{figure}

Figure~\ref{fig:ShovelAttacks} presents the outputs of FGSM, BIM, and our universal perturbation attacks, respectively, all applied with the same noise level, $\epsilon=0.05$, to five different viewing angles of the shovel. Each training image displays the predicted label probability for the correct class. Before the attacks, these training images were correctly classified with confidence ranging from 0.64 to 0.88. 

\section{Discussion and Limitations}\label{sec:Discussion}




In this paper, the universal attacks do not target a user-specified label. A least-likely penalty, as proposed in ILCM, could be added to the loss function to specify a desired adversarial target. When extending this method to targeted attacks, we suspect that a larger number of iterations may then be necessary. For untargeted attacks presented here, we found a single epoch to be sufficient. We leave targeted attacks to be future work. 


Another potential difference between Universal and FGSM/BIM lies in the initialization of the adversarial noise. Our method may be sensitive to the initialization of adversarial noise, unlike BIM or FGSM methods. Since we calculate the gradients with respect to the noise, instead of the image, certain initial values of noise can cause numerical instabilities, including initializing noise to zero. Consequently, initial adversarial noise could be considered another parameter to be tuned in the future. Currently, we initialize the noise as a uniformly distributed value between $\{-0.01, 0.01\}$. Future endeavors may involve developing an algorithm to determine the optimal initialization or providing a default initialization that is effective across different objects.


While further work is needed to address these points, our method offers distinct advantages like efficiency, multi-view applicability, and potential scalability. Future research could explore techniques to enhance accuracy, investigate targeted attack scenarios, and incorporate different classification models for broader applicability.




\section{Conclusion}\label{sec:conclusion}




In this work, we presented a novel ``universal perturbation'' method for generating robust multi-view adversarial examples for 3D object recognition. Our method operates solely on 2D images, offering a practical and scalable alternative to computationally expensive 3D adversarial attacks. By crafting a single noise perturbation applicable to various object views, our approach achieves 3D-like attack capabilities while retaining the efficiency and practicality of single-view methods.

 Experiments on diverse 3D objects demonstrated the effectiveness of our approach. Our universal attacks successfully identified single noise perturbations with higher destruction rates across multiple viewing angles, especially at low $\epsilon$ levels. This performance highlights the improvements over conventional single-view attacks, which may struggle with viewpoint variations or need to identify a noise for each unique viewpoint.

 Beyond its effectiveness, our method offers key advantages in terms of efficiency and scalability. By requiring only a single perturbation generation instead of per-view computations, our approach reduces computational resources compared to other methods. This efficiency makes it particularly well-suited for real-world applications where computational constraints or multi-view points are present.

 While our work represents a significant step forward, we acknowledge potential limitations and opportunities for future research. Further evaluation of a wider range of object categories and classification models could enhance the generalizability of our findings. Additionally, extending our method for targeted attacks is an intriguing direction for future work.


%
%

\bibliographystyle{ACM-Reference-Format}
\bibliography{refs}


\begin{thebibliography}{30}


\ifx \showCODEN    \undefined \def \showCODEN     #1{\unskip}     \fi
\ifx \showDOI      \undefined \def \showDOI       #1{#1}\fi
\ifx \showISBNx    \undefined \def \showISBNx     #1{\unskip}     \fi
\ifx \showISBNxiii \undefined \def \showISBNxiii  #1{\unskip}     \fi
\ifx \showISSN     \undefined \def \showISSN      #1{\unskip}     \fi
\ifx \showLCCN     \undefined \def \showLCCN      #1{\unskip}     \fi
\ifx \shownote     \undefined \def \shownote      #1{#1}          \fi
\ifx \showarticletitle \undefined \def \showarticletitle #1{#1}   \fi
\ifx \showURL      \undefined \def \showURL       {\relax}        \fi
\providecommand\bibfield[2]{#2}
\providecommand\bibinfo[2]{#2}
\providecommand\natexlab[1]{#1}
\providecommand\showeprint[2][]{arXiv:#2}

\bibitem[Andriushchenko and Flammarion(2020)]%
        {andriushchenko2020understanding}
\bibfield{author}{\bibinfo{person}{Maksym Andriushchenko} {and} \bibinfo{person}{Nicolas Flammarion}.} \bibinfo{year}{2020}\natexlab{}.
\newblock \showarticletitle{Understanding and improving fast adversarial training}.
\newblock \bibinfo{journal}{\emph{Advances in Neural Information Processing Systems}}  \bibinfo{volume}{33} (\bibinfo{year}{2020}), \bibinfo{pages}{16048--16059}.
\newblock


\bibitem[Apicharttrisorn et~al\mbox{.}(2019)]%
        {apicharttrisorn2019frugal}
\bibfield{author}{\bibinfo{person}{Kittipat Apicharttrisorn}, \bibinfo{person}{Xukan Ran}, \bibinfo{person}{Jiasi Chen}, \bibinfo{person}{Srikanth~V Krishnamurthy}, {and} \bibinfo{person}{Amit~K Roy-Chowdhury}.} \bibinfo{year}{2019}\natexlab{}.
\newblock \showarticletitle{Frugal following: Power thrifty object detection and tracking for mobile augmented reality}. In \bibinfo{booktitle}{\emph{Proceedings of the 17th Conference on Embedded Networked Sensor Systems}}. \bibinfo{pages}{96--109}.
\newblock


\bibitem[Bes(2017)]%
        {baseball_cite}
\bibfield{author}{\bibinfo{person}{Alex Bes}.} \bibinfo{year}{2017}\natexlab{}.
\newblock \bibinfo{title}{Worn Baseball Ball}.
\newblock
\newblock
\urldef\tempurl%
\url{https://sketchfab.com/3d-models/worn-baseball-ball-fdf3de6ae225421ea78961b897b9608a}
\showURL{%
\tempurl}
\newblock
\shownote{Last accessed 10 February 2024}.


\bibitem[Carlini and Wagner(2017)]%
        {carlini2017evaluating}
\bibfield{author}{\bibinfo{person}{Nicholas Carlini} {and} \bibinfo{person}{David Wagner}.} \bibinfo{year}{2017}\natexlab{}.
\newblock \bibinfo{title}{Towards Evaluating the Robustness of Neural Networks}.
\newblock
\newblock
\showeprint[arxiv]{1608.04644}~[cs.CR]


\bibitem[dannyboy70000(2014)]%
        {lemon_cite}
\bibfield{author}{\bibinfo{person}{dannyboy70000}.} \bibinfo{year}{2014}\natexlab{}.
\newblock \bibinfo{title}{lemon 3D Model}.
\newblock
\newblock
\urldef\tempurl%
\url{https://free3d.com/3d-model/lemon-72357.html}
\showURL{%
\tempurl}
\newblock
\shownote{Last accessed 10 February 2024}.


\bibitem[Deng et~al\mbox{.}(2009)]%
        {deng2009imagenet}
\bibfield{author}{\bibinfo{person}{Jia Deng}, \bibinfo{person}{Wei Dong}, \bibinfo{person}{Richard Socher}, \bibinfo{person}{Li-Jia Li}, \bibinfo{person}{Kai Li}, {and} \bibinfo{person}{Li Fei-Fei}.} \bibinfo{year}{2009}\natexlab{}.
\newblock \showarticletitle{Imagenet: A large-scale hierarchical image database}. In \bibinfo{booktitle}{\emph{2009 IEEE conference on computer vision and pattern recognition}}. Ieee, \bibinfo{pages}{248--255}.
\newblock


\bibitem[Dong et~al\mbox{.}(2018)]%
        {dong2018boosting}
\bibfield{author}{\bibinfo{person}{Yinpeng Dong}, \bibinfo{person}{Fangzhou Liao}, \bibinfo{person}{Tianyu Pang}, \bibinfo{person}{Hang Su}, \bibinfo{person}{Jun Zhu}, \bibinfo{person}{Xiaolin Hu}, {and} \bibinfo{person}{Jianguo Li}.} \bibinfo{year}{2018}\natexlab{}.
\newblock \bibinfo{title}{Boosting Adversarial Attacks with Momentum}.
\newblock
\newblock
\showeprint[arxiv]{1710.06081}~[cs.LG]


\bibitem[Finn et~al\mbox{.}(2015)]%
        {finn2015learning}
\bibfield{author}{\bibinfo{person}{Chelsea Finn}, \bibinfo{person}{Xin~Yu Tan}, \bibinfo{person}{Yan Duan}, \bibinfo{person}{Trevor Darrell}, \bibinfo{person}{Sergey Levine}, {and} \bibinfo{person}{Pieter Abbeel}.} \bibinfo{year}{2015}\natexlab{}.
\newblock \showarticletitle{Learning visual feature spaces for robotic manipulation with deep spatial autoencoders}.
\newblock \bibinfo{journal}{\emph{arXiv preprint arXiv:1509.06113}}  \bibinfo{volume}{25} (\bibinfo{year}{2015}), \bibinfo{pages}{2}.
\newblock


\bibitem[GetDeadEntertainment(2020)]%
        {shovel_cite}
\bibfield{author}{\bibinfo{person}{GetDeadEntertainment}.} \bibinfo{year}{2020}\natexlab{}.
\newblock \bibinfo{title}{Medieval Shovel}.
\newblock
\newblock
\urldef\tempurl%
\url{https://www.turbosquid.com/3d-models/medieval-shovel-3d-model-1494436}
\showURL{%
\tempurl}
\newblock
\shownote{Last accessed 10 February 2024}.


\bibitem[Goodfellow et~al\mbox{.}(2015)]%
        {goodfellow2015explaining}
\bibfield{author}{\bibinfo{person}{Ian~J. Goodfellow}, \bibinfo{person}{Jonathon Shlens}, {and} \bibinfo{person}{Christian Szegedy}.} \bibinfo{year}{2015}\natexlab{}.
\newblock \bibinfo{title}{Explaining and Harnessing Adversarial Examples}.
\newblock
\newblock
\showeprint[arxiv]{1412.6572}~[stat.ML]


\bibitem[Hossain et~al\mbox{.}(2016)]%
        {hossain2016object}
\bibfield{author}{\bibinfo{person}{Delowar Hossain}, \bibinfo{person}{Genci Capi}, {and} \bibinfo{person}{Mitsuru Jindai}.} \bibinfo{year}{2016}\natexlab{}.
\newblock \showarticletitle{Object recognition and robot grasping: A deep learning based approach}. In \bibinfo{booktitle}{\emph{The 34th Annual Conference of the Robotics Society of Japan (RSJ 2016), Yamagata, Japan}}.
\newblock


\bibitem[Hu et~al\mbox{.}(2019)]%
        {hu20193}
\bibfield{author}{\bibinfo{person}{Zhe Hu}, \bibinfo{person}{Tao Han}, \bibinfo{person}{Peigen Sun}, \bibinfo{person}{Jia Pan}, {and} \bibinfo{person}{Dinesh Manocha}.} \bibinfo{year}{2019}\natexlab{}.
\newblock \showarticletitle{3-D deformable object manipulation using deep neural networks}.
\newblock \bibinfo{journal}{\emph{IEEE Robotics and Automation Letters}} \bibinfo{volume}{4}, \bibinfo{number}{4} (\bibinfo{year}{2019}), \bibinfo{pages}{4255--4261}.
\newblock


\bibitem[Klokov and Lempitsky(2017)]%
        {klokov2017escape}
\bibfield{author}{\bibinfo{person}{Roman Klokov} {and} \bibinfo{person}{Victor Lempitsky}.} \bibinfo{year}{2017}\natexlab{}.
\newblock \showarticletitle{Escape from cells: Deep kd-networks for the recognition of 3d point cloud models}. In \bibinfo{booktitle}{\emph{Proceedings of the IEEE international conference on computer vision}}. \bibinfo{pages}{863--872}.
\newblock


\bibitem[Kurakin et~al\mbox{.}(2017)]%
        {kurakin2017adversarial}
\bibfield{author}{\bibinfo{person}{Alexey Kurakin}, \bibinfo{person}{Ian Goodfellow}, {and} \bibinfo{person}{Samy Bengio}.} \bibinfo{year}{2017}\natexlab{}.
\newblock \bibinfo{title}{Adversarial examples in the physical world}.
\newblock
\newblock
\showeprint[arxiv]{1607.02533}~[cs.CV]


\bibitem[Le and Duan(2018)]%
        {le2018pointgrid}
\bibfield{author}{\bibinfo{person}{Truc Le} {and} \bibinfo{person}{Ye Duan}.} \bibinfo{year}{2018}\natexlab{}.
\newblock \showarticletitle{Pointgrid: A deep network for 3d shape understanding}. In \bibinfo{booktitle}{\emph{Proceedings of the IEEE conference on computer vision and pattern recognition}}. \bibinfo{pages}{9204--9214}.
\newblock


\bibitem[Li et~al\mbox{.}(2019)]%
        {li2019stereo}
\bibfield{author}{\bibinfo{person}{Peiliang Li}, \bibinfo{person}{Xiaozhi Chen}, {and} \bibinfo{person}{Shaojie Shen}.} \bibinfo{year}{2019}\natexlab{}.
\newblock \showarticletitle{Stereo r-cnn based 3d object detection for autonomous driving}. In \bibinfo{booktitle}{\emph{Proceedings of the IEEE/CVF Conference on Computer Vision and Pattern Recognition}}. \bibinfo{pages}{7644--7652}.
\newblock


\bibitem[Li et~al\mbox{.}(2020)]%
        {li2020object}
\bibfield{author}{\bibinfo{person}{Xiang Li}, \bibinfo{person}{Yuan Tian}, \bibinfo{person}{Fuyao Zhang}, \bibinfo{person}{Shuxue Quan}, {and} \bibinfo{person}{Yi Xu}.} \bibinfo{year}{2020}\natexlab{}.
\newblock \showarticletitle{Object detection in the context of mobile augmented reality}. In \bibinfo{booktitle}{\emph{2020 IEEE International Symposium on Mixed and Augmented Reality (ISMAR)}}. IEEE, \bibinfo{pages}{156--163}.
\newblock


\bibitem[Lin et~al\mbox{.}(2020)]%
        {lin2020nesterov}
\bibfield{author}{\bibinfo{person}{Jiadong Lin}, \bibinfo{person}{Chuanbiao Song}, \bibinfo{person}{Kun He}, \bibinfo{person}{Liwei Wang}, {and} \bibinfo{person}{John~E. Hopcroft}.} \bibinfo{year}{2020}\natexlab{}.
\newblock \bibinfo{title}{Nesterov Accelerated Gradient and Scale Invariance for Adversarial Attacks}.
\newblock
\newblock
\showeprint[arxiv]{1908.06281}~[cs.LG]


\bibitem[Liu et~al\mbox{.}(2019)]%
        {liu2019edge}
\bibfield{author}{\bibinfo{person}{Luyang Liu}, \bibinfo{person}{Hongyu Li}, {and} \bibinfo{person}{Marco Gruteser}.} \bibinfo{year}{2019}\natexlab{}.
\newblock \showarticletitle{Edge assisted real-time object detection for mobile augmented reality}. In \bibinfo{booktitle}{\emph{The 25th annual international conference on mobile computing and networking}}. \bibinfo{pages}{1--16}.
\newblock


\bibitem[Luo et~al\mbox{.}(2022)]%
        {luo2022entangling}
\bibfield{author}{\bibinfo{person}{Xi-Yu Luo}, \bibinfo{person}{Yong Yu}, \bibinfo{person}{Jian-Long Liu}, \bibinfo{person}{Ming-Yang Zheng}, \bibinfo{person}{Chao-Yang Wang}, \bibinfo{person}{Bin Wang}, \bibinfo{person}{Jun Li}, \bibinfo{person}{Xiao Jiang}, \bibinfo{person}{Xiu-Ping Xie}, \bibinfo{person}{Qiang Zhang}, {et~al\mbox{.}}} \bibinfo{year}{2022}\natexlab{}.
\newblock \showarticletitle{Entangling metropolitan-distance separated quantum memories}.
\newblock \bibinfo{journal}{\emph{arXiv preprint arXiv:2201.11953}} (\bibinfo{year}{2022}).
\newblock


\bibitem[Madry et~al\mbox{.}(2017)]%
        {madry2017towards}
\bibfield{author}{\bibinfo{person}{Aleksander Madry}, \bibinfo{person}{Aleksandar Makelov}, \bibinfo{person}{Ludwig Schmidt}, \bibinfo{person}{Dimitris Tsipras}, {and} \bibinfo{person}{Adrian Vladu}.} \bibinfo{year}{2017}\natexlab{}.
\newblock \showarticletitle{Towards deep learning models resistant to adversarial attacks}.
\newblock \bibinfo{journal}{\emph{arXiv preprint arXiv:1706.06083}} (\bibinfo{year}{2017}).
\newblock


\bibitem[mitsui(2020)]%
        {table_cite}
\bibfield{author}{\bibinfo{person}{mitsui}.} \bibinfo{year}{2020}\natexlab{}.
\newblock \bibinfo{title}{table 3D Model}.
\newblock
\newblock
\urldef\tempurl%
\url{https://free3d.com/3d-model/table-747735.html}
\showURL{%
\tempurl}
\newblock
\shownote{Last accessed 10 February 2024}.


\bibitem[Qi et~al\mbox{.}(2017)]%
        {qi2017pointnet}
\bibfield{author}{\bibinfo{person}{Charles~R Qi}, \bibinfo{person}{Hao Su}, \bibinfo{person}{Kaichun Mo}, {and} \bibinfo{person}{Leonidas~J Guibas}.} \bibinfo{year}{2017}\natexlab{}.
\newblock \showarticletitle{Pointnet: Deep learning on point sets for 3d classification and segmentation}. In \bibinfo{booktitle}{\emph{Proceedings of the IEEE conference on computer vision and pattern recognition}}. \bibinfo{pages}{652--660}.
\newblock


\bibitem[Ravindran et~al\mbox{.}(2020)]%
        {ravindran2020multi}
\bibfield{author}{\bibinfo{person}{Ratheesh Ravindran}, \bibinfo{person}{Michael~J Santora}, {and} \bibinfo{person}{Mohsin~M Jamali}.} \bibinfo{year}{2020}\natexlab{}.
\newblock \showarticletitle{Multi-object detection and tracking, based on DNN, for autonomous vehicles: A review}.
\newblock \bibinfo{journal}{\emph{IEEE Sensors Journal}} \bibinfo{volume}{21}, \bibinfo{number}{5} (\bibinfo{year}{2020}), \bibinfo{pages}{5668--5677}.
\newblock


\bibitem[Ren et~al\mbox{.}(2020)]%
        {ren2020adversarial}
\bibfield{author}{\bibinfo{person}{Kui Ren}, \bibinfo{person}{Tianhang Zheng}, \bibinfo{person}{Zhan Qin}, {and} \bibinfo{person}{Xue Liu}.} \bibinfo{year}{2020}\natexlab{}.
\newblock \showarticletitle{Adversarial attacks and defenses in deep learning}.
\newblock \bibinfo{journal}{\emph{Engineering}} \bibinfo{volume}{6}, \bibinfo{number}{3} (\bibinfo{year}{2020}), \bibinfo{pages}{346--360}.
\newblock


\bibitem[selfie 3D~scan(2019)]%
        {tractor_cite}
\bibfield{author}{\bibinfo{person}{selfie 3D~scan}.} \bibinfo{year}{2019}\natexlab{}.
\newblock \bibinfo{title}{Tractor}.
\newblock
\newblock
\urldef\tempurl%
\url{https://sketchfab.com/3d-models/tractor-1b258bcc01bf4ed0935ef73e80442c30}
\showURL{%
\tempurl}
\newblock
\shownote{Last accessed 10 February 2024}.


\bibitem[Szegedy et~al\mbox{.}(2013)]%
        {szegedy2013intriguing}
\bibfield{author}{\bibinfo{person}{Christian Szegedy}, \bibinfo{person}{Wojciech Zaremba}, \bibinfo{person}{Ilya Sutskever}, \bibinfo{person}{Joan Bruna}, \bibinfo{person}{Dumitru Erhan}, \bibinfo{person}{Ian Goodfellow}, {and} \bibinfo{person}{Rob Fergus}.} \bibinfo{year}{2013}\natexlab{}.
\newblock \showarticletitle{Intriguing properties of neural networks}.
\newblock \bibinfo{journal}{\emph{arXiv preprint arXiv:1312.6199}} (\bibinfo{year}{2013}).
\newblock


\bibitem[Wang et~al\mbox{.}(2019)]%
        {wang2019normalnet}
\bibfield{author}{\bibinfo{person}{Cheng Wang}, \bibinfo{person}{Ming Cheng}, \bibinfo{person}{Ferdous Sohel}, \bibinfo{person}{Mohammed Bennamoun}, {and} \bibinfo{person}{Jonathan Li}.} \bibinfo{year}{2019}\natexlab{}.
\newblock \showarticletitle{NormalNet: A voxel-based CNN for 3D object classification and retrieval}.
\newblock \bibinfo{journal}{\emph{Neurocomputing}}  \bibinfo{volume}{323} (\bibinfo{year}{2019}), \bibinfo{pages}{139--147}.
\newblock


\bibitem[Zhang et~al\mbox{.}(2020)]%
        {zhang2020dnn}
\bibfield{author}{\bibinfo{person}{K Zhang}, \bibinfo{person}{SJ Wang}, \bibinfo{person}{L Ji}, {and} \bibinfo{person}{C Wang}.} \bibinfo{year}{2020}\natexlab{}.
\newblock \showarticletitle{DNN based camera and LiDAR fusion framework for 3D object recognition}. In \bibinfo{booktitle}{\emph{Journal of Physics: Conference Series}}, Vol.~\bibinfo{volume}{1518}. IOP Publishing, \bibinfo{pages}{012044}.
\newblock


\bibitem[Zhi et~al\mbox{.}(2018)]%
        {zhi2018toward}
\bibfield{author}{\bibinfo{person}{Shuaifeng Zhi}, \bibinfo{person}{Yongxiang Liu}, \bibinfo{person}{Xiang Li}, {and} \bibinfo{person}{Yulan Guo}.} \bibinfo{year}{2018}\natexlab{}.
\newblock \showarticletitle{Toward real-time 3D object recognition: A lightweight volumetric CNN framework using multitask learning}.
\newblock \bibinfo{journal}{\emph{Computers \& Graphics}}  \bibinfo{volume}{71} (\bibinfo{year}{2018}), \bibinfo{pages}{199--207}.
\newblock


\end{thebibliography}
%





\end{document}